# Top-Down Unsupervised Image Segmentation
# (it sounds like an oxymoron, but actually it isn't)

Emanuel Diamant
VIDIA-mant, POB 933, Kiriat Ono 55100, Israel
(emanl@012.net.il)

*Abstract* – Pattern recognition is generally assumed as an interaction of two inversely directed image-processing streams: the bottom-up information details gathering and localization (segmentation) stream, and the top-down information features aggregation, association and interpretation (recognition) stream. Inspired by recent evidence from biological vision research and by the insights of Kolmogorov Complexity theory, we propose a new, just top-down evolving, procedure of initial image segmentation. We claim that traditional top-down cognitive reasoning, which is supposed to guide the segmentation process to its final result, is not at all a part of the image information content evaluation. And that initial image segmentation is certainly an unsupervised process. We present some illustrative examples, which support our claims.

## 1. INTRODUCTION

Pattern recognition is generally approached as an interaction of two inversely proceeding image-processing streams: the bottom-up information features localization and segmentation procedure, and the top-down information features association, aggregation and interpretation (recognition) procedure.

The roots of such an approach can be traced back to Treisman's Feature Integrating Theory [1] or Biederman's Recognition-by-components theory [2]. Biological vision has always been an unlimited source of inspiration for computer vision (systems) designers. It was only natural that when computer vision society had matured enough to cope with pattern recognition problems, appropriate biological vision solutions have been accepted and adopted by the computer vision society and have become an indispensable part of computer vision practice [3], [4].

However, straightforward imitation of biological vision solutions is not always possible. For example, how the knowledge about information features interrelations (features' semantics, usually acquired in course of learning and natural experience of the living beings) is made available to the reasoning stage of a computer vision system? How this external knowledge is incorporated into the segmentation and pattern identification processes?

## 2. STATE OF THE ART

Contemporary computer vision systems don't have clear answers to these questions. Some of them rely on expert knowledge (squeezed and restricted to a specific application) which is imprinted into the system at the design time. Some attempt to provide limited initial training and concept-restricted adaptation facilities. However, the basic principle of a bottom-up initial information content gathering remains invariable.

According to this principle, low-level elementary information pieces are first searched and evaluated over the entire image space. Then they are grouped and aggregated into larger and more complex agglomerations, which are fed to the higher system levels for farther generalization, classification, and other higher-level processing [5]. To accommodate for external (user or system) requirements, that is, to incorporate the rules and principles by which disordered information pieces are combined and agglomerated, a top-down control flow is generally assumed. Its aim is to mediate the bottom-up information gathering. It is generally believed that this supervised intervention of a top-down conscious control will lead to a more suitable and more task-fitting low-level information features acquisition [6], [7].

There is no need to say that such an approach, where every pixel within an image must be visited and investigated at the preprocessing stage, can not faithfully serve the tough requirements of the modern vision systems. It is also hard to believe that in course of natural evolution such an awkward and inefficient mechanism has been survived and sustained the natural selection.

## 3. A DIFFERENT APPROACH

New evidence from biological vision research, which became available in the recent years, put the correctness of the traditional approach in doubts. To properly understand the point, a couple of words must be spent on human vision peculiarities: Human eye's retina is a bizarre structure – only a small fraction of its view field (approximately $2^0$ out of the entire field of $160^0$, [8]) is densely populated with photoreceptors. Just this small fragment of the retina (the so-called fovea) is responsible for our ability to see a sharp and clear picture of the surrounding world. The rest of the view field is a fast descending (in spatial density) placement of light sensitive photoreceptors. To compensate for the lack of resolution over the entire field, continuous eye movements (also known as eye saccades) are performed, sequentially placing the high-resolution fovea over various scene locations.

According to attention vision theories, the decision to make a saccade and to fix the fovea over a new image location *precedes* the high-resolution (low-level) image information gathering, and hence, it can be yielded only by the coarse and poor information delivered from the peripheral vision areas. The flow of new evidence convincingly supports this unusual hypothesis: visual recognition/categorization tasks use "express", but comparatively imprecise and coarse-scale representations, before the fine-scale representations are acquired [9]; the first signals reaching the highest processing levels are from the eye's periphery, not from the fovea [10]. Not less surprising is the evidence that traditional assumptions about top-down intervention from the upper cognitive levels simply do not hold here. In most of the cases, saccadic movements are guided preattentively and unconsciously. In computer vision terminology that means that initial information gathering is definitely a top-down process, but the term "top-down" has here a quite different interpretation. The new top-down segmentation procedure definitely commences with a squeezed and simplified image representation (delivered by the eye's periphery) and the required information details are unveiled gradually at the lower (fovea closer, information rich) representation levels [11].

Underpinning theoretical formalization of the observed biological evidence can be deduced from the Kolmogorov Complexity theory [12]. Originally devised to explore the notion of randomness, it has introduced a new definition of information content. Contrary to the well-known and ubiquitously exploited Shannon's definition, which can be seen as an average information (average randomness) over the whole ensemble, Kolmogorov complexity allows to deal with the quantity of information in individual objects. We briefly summarize the principal claims of Kolmogorov complexity theory (slightly twisted to fit the case of image information content exploration).

- Image information content is a set of descriptions of the observable image data structures.
- These descriptions are executable, that is, following them the meaningful part of image content can be faithfully reconstructed.
- These descriptions are hierarchical and recursive, that is, starting with a generalized and simplified description of image structure they proceed in a top-down fashion to more and more fine information details resolved at lower description levels.
- Although the lower bound of description details is unattainable, that does not pose a serious problem because information content comprehension is generally fine details devoid.

More rigorous presentation of these subjects can be find in numerous theme-related publications [12 – 14].

### 4. IMPLEMENTATION

Following the modern concepts of selective attention vision and the insights of Kolmogorov Complexity theory, we dare to propose a new scheme for information content gathering. . Its architecture is shown in Figure 1, and it is comprised of three main processing paths: the bottom-up processing path, the top-down processing path and a stack where the discovered information content (the generated descriptions of it) are actually accumulated.

To facilitate the requirement for a top-down directed processing, we introduce a hierarchy of multi-level multi-resolution image representations called multi-stage image pyramid [15]. Such pyramid construction generates a set of compressed copies of the original input image. Each image in the sequence can be seen as an array each dimension of which is half as large as its predecessor. The rules of this shrinking operation are very simple and fast: four non-overlapping neighbour pixels in an image at level $L$ are averaged and the result is assigned to a pixel in a higher ($L$+1)-level image. This is known as "four children to one parent relationship".

At the top of the pyramid, the resulting coarse image undergoes a round of further simplification. Several image zones, representing perceptually discernible image fractions (visually dominated image parts, super-objects) are determined (segmented) and identified by assigning labels to each of the segmented pieces. Since the image size at the top is significantly reduced and since in the course of the bottom-up image squeezing a severe data averaging is attained, the image segmentation/classification procedure does not demand special computational resources. Thus, any well-known segmentation methodology will suffice. We use our own proprietary technique that is based on a low-level (local) information content evaluation [16].

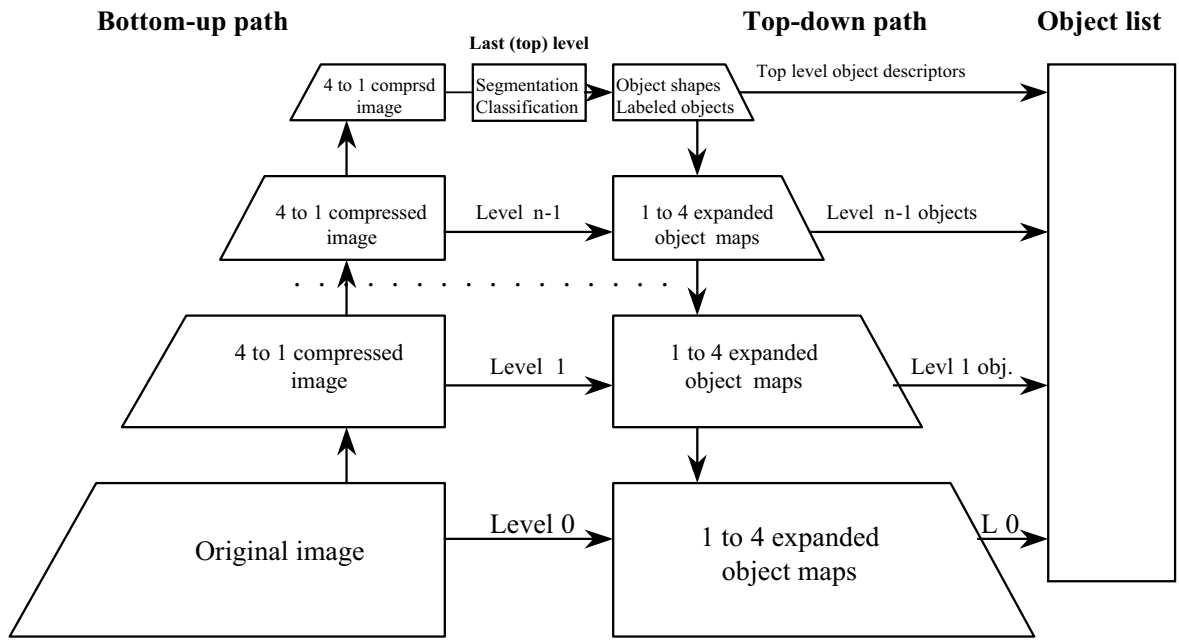

Fig. 1. The Block Diagram (Schema) of the proposed approach

This technique first outlines the borders of the principal image fragments. Then similarly appearing pixels within the borders are aggregated in compact spatially connected regional groups (clusters). Afterwards, every cluster is marked with a label. Thus, a map of labeled clusters, corresponding to perceptually discernible image regions, is produced. Finally, to accomplish top-level object identification, for each labeled region its characteristic intensity is computed as an average of labeled pixels. This way, a second (additional) segmentation map is produced, where regions are represented by their characteristic intensities.

From this point on, the top-down processing path is commenced. At each level, the two previously defined maps are expanded to the size of the image at the nearest lower level. The expansion rule is very simple: the value of each parent pixel is assigned to its four children in the corresponding lower-level map (a reversed shrinking operation). Since the regions at different hierarchical levels do not exhibit significant changes in their characteristic intensity, the majority of newly assigned pixels are determined in a sufficiently correct manner. Only pixels at region borders (and seeds of newly emerging regions) may significantly deviate from the assigned values. Taking the corresponding current-level image as a reference (the left side, bottom-up path belonging images), these pixels can be easily detected and subjected to a refinement cycle. Here they are allowed to adjust themselves to the "proper" nearest neighbors, which certainly belong to one of the previously labeled regions (or to the newly emerging ones).

In such a manner, the process is subsequently repeated at all descending levels until the segmentation/classification of the zero-level (original input image) is successfully accomplished. At every processing level, every image object/region (just recovered or an inherited one) is registered in the objects' appearance list, which is the third constituting part of the proposed scheme. The registered object parameters are the available simplified object's attributes, such as size, center-of-mass position, average object intensity and hierarchical and topological relationship within and between the objects ("sub-part of…", "at the left of…", etc.). They are sparse, general, and yet specific enough to capture the object's characteristic features in a variety of descriptive forms.

## 5. EXPERIMENTAL RESULTS

To illustrate the qualities of the proposed approach we have chosen an Aerial View of the Lupa Center from the NASA Grin Collection [17].

Fig. 2 represents the original image, Figs. 3, 4, and 5 demonstrate how the original image is decomposed by the proposed routine to regions of various detail description complexity. Level 5 (Fig. 3) corresponds to the near the top-most hierarchical level (this particular image results in a 6-level hierarchy). Level 1 (Fig. 5) is the lower end closest decomposition.

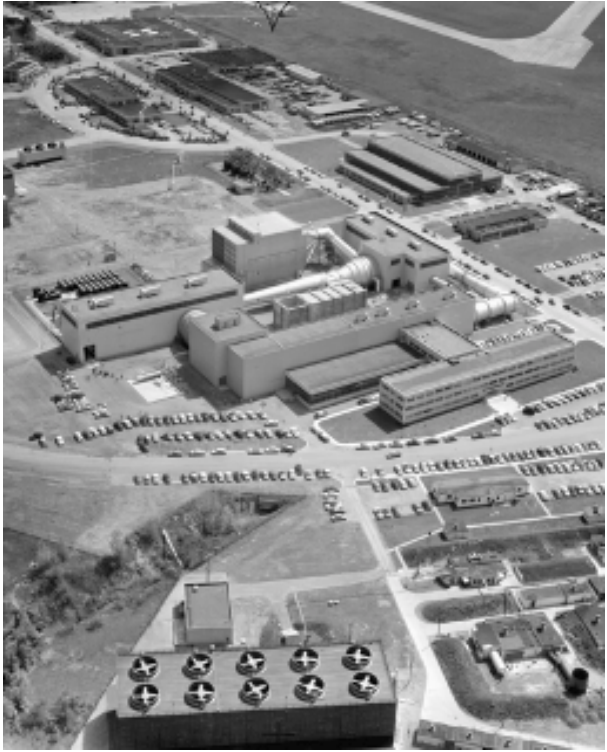

Fig. 2. Original image, size 512 x 640 pixels.

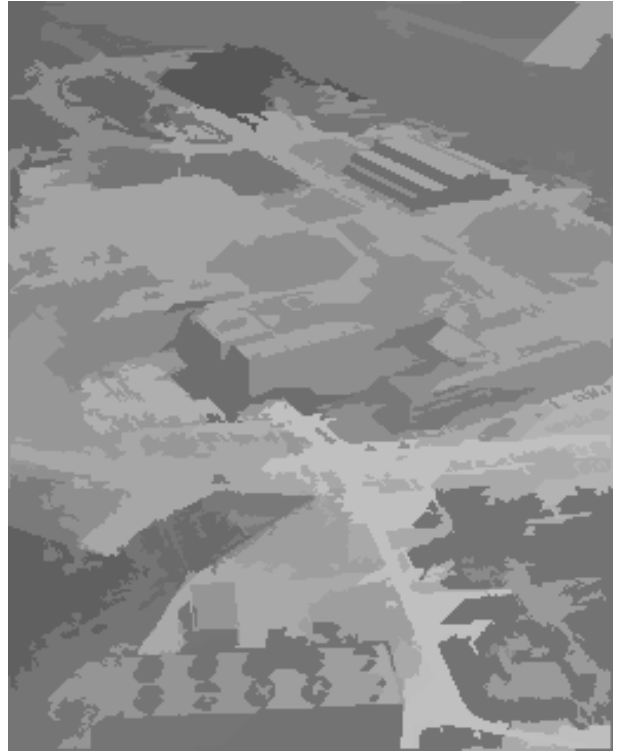

Fig. 3. Level 5 segmentation, 32 regions.

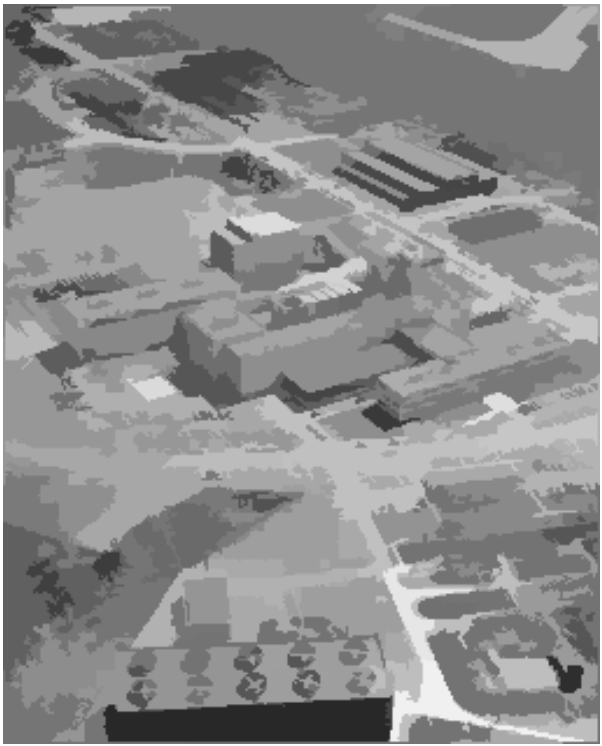

Fig. 4. Level 3 segmentation, 108 regions.

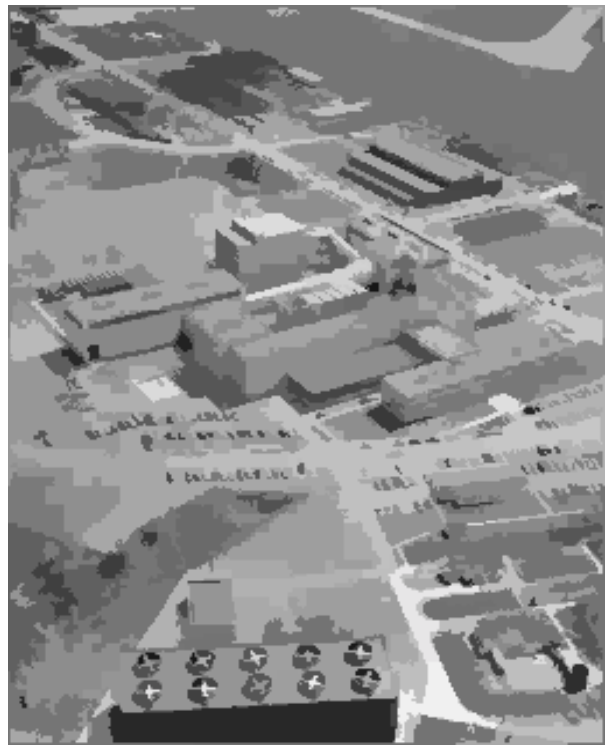

Fig. 5. Level 1 segmentation, 235 regions.

For space saving, we provide only few examples from the image segmentation gallery, which for the reader's convenience are all expanded to the original image size. Extracted from the object list, the numbers of distinguished (segmented) at each corresponding level regions are also given in each figure capture.

Because in our approach the real target objects are not known in advance, only the generalized intensity maps (of the possible regions) are presented. Every one can see that the simplified representation is sufficient to grasp the image concept, that is, to interpret the provided image information content. It is now easy for the user to define what region combination depicts the target object most faithfully, and then, using the rest of the available descriptions (not shown here, but contained in the object list), to make a comparison analysis for successful target identification.

## 6. CONCLUSIONS

We have proposed a new region-based information content description approach, which solely relies on a top-down processing methodology. Contrary to traditional approaches, which rely on a bottom-up (resource exhaustive) processing and on a top-down mediating intervention (intended for user's external knowledge incorporation), our proposed schema is an entirely top-down evolving procedure (that means, considerable computational load shrinking is attained), which is indifferent to any user or task-related requirements.

Ubiquitous claims for "unsupervised" or "automatic" image segmentation techniques must be modestly put aside. In spite of the fact, that our top-down region decomposition (segmentation) is definitely user independent and unsupervised, the final decision about the segmented region importance is an indisputable user's prerogative. Contrary to traditional approaches, we put it external to the segmentation path. At each segmentation level, the resulting region descriptions (from the object list, which is a part of the scheme) are available to the user (or to the system's decision-making unit) for further analysis, comparison, and final object (region, pattern) recognition (classification, determination, interpretation, etc.).